\documentclass[conference]{IEEEtran}
\IEEEoverridecommandlockouts

\usepackage{amsfonts, amssymb, mathtools, amsmath, amsthm}
\usepackage{cite} 
\usepackage{multicol}
\usepackage{tikz}
\usepackage{acronym} 
\usepackage{subcaption} 
\usetikzlibrary{automata, positioning, arrows} 
\usepackage{graphicx} 
\usepackage{arydshln} 
\usepackage{algorithm} 
\usepackage{algorithmic} 
\usepackage{pgfplots}  
\pgfplotsset{compat=1.16}
\usepackage{xcolor} 
\usepackage[bookmarks=true,hidelinks,pdftitle={Topology-Guided Modular Actor-Critic Learning for Continuous Systems under Temporal Objectives},pdfauthor={Lening Li, Zhentian Qian, Jianan Xia, Yawen Wang, Zhongjing Li, Qiren Geng, Huasheng Zhang, Liang Hu, Qishuang Li, Junqiang Lou}]{hyperref} 
\usepackage{orcidlink} 
\usepackage{url}
\newif\ifuseboldmathops
\newif\ifuseittextabbrevs
\useboldmathopstrue   

\ifuseittextabbrevs
    
    \newcommand{\eg}{{\it e.g.}}
    \newcommand{\ie}{{\it i.e.}}

\else
    
    \newcommand{\eg}{e.g.}
    \newcommand{\ie}{i.e.}

\fi

\ifuseboldmathops
    \newcommand{\reals}{\mathbb{R}}

\else
    \newcommand{\reals}{\mathbb{R}}

\fi

\ifuseboldmathops

\else

\fi

\ifuseboldmathops
    \newcommand{\Expect}{\mathop{\bf E{}}\nolimits}
    \newcommand{\Prob}{\mathop{\bf P{}}\nolimits}
\else
    \newcommand{\Expect}{\mathop{\mathbb{E}{}}\nolimits}
    \newcommand{\Prob}{\mathop{\mathbb{P}{}}\nolimits}
\fi

\ifuseboldmathops

\else

\fi

\newcommand{\argmax}{\mathop{\mathrm{argmax}}}

\newcommand{\Always}{\Box \, }
\newcommand{\Eventually}{\Diamond \, }
\newcommand{\Next}{\bigcirc \, }
\newcommand{\until}{\mbox{$\, {\sf U}\,$}}

\newcommand{\Lagrange}{\mathcal{L}}

\newcommand{\sink}{\mathsf{sink}}

\newcommand{\norm}[1]{\lVert#1\rVert}

\newcommand{\abs}[1]{\lvert#1\rvert}

\newcommand{\calD}{\mathcal{D}}
\newcommand{\subgoal}{\mathsf{subgoal}}

\DeclareMathOperator{\guard}{Guard}

\DeclareMathOperator{\level}{\mathcal{L}}
\DeclareMathOperator{\partition}{\mathcal{X}}

\newcommand{\dfa}{\mathcal{A}}

\newcommand{\calAP}{\mathcal{AP}}

\newcommand{\indicator}{\mathbf{1}}

\newcommand{\relation}{\rightharpoonup}
\newcommand{\brelation}{\rightleftharpoons}
\newcommand{\topo}{\rightsquigarrow}

\DeclareMathOperator{\mdp}{M}
\DeclareMathOperator{\env}{\mathsf{Env}}
\DeclareMathOperator{\mellowmax}{\mathcal{T}}
\renewcommand{\path}{\mathsf{path}}

\DeclareMathOperator{\prodmdp}{\mathcal{M}}
\DeclareMathOperator{\prodtrans}{\Delta}
\DeclareMathOperator{\prodf}{\mathcal{F}}

\newcommand{\pr}{\mbox{Pr}}

\newcommand{\val}{\mathcal{V}}
\newcommand{\qval}{\mathcal{Q}}

\theoremstyle{definition}
\newtheorem{definition}{Definition}
\newtheorem{example}{Example}
\newtheorem{theorem}{Theorem}

\newtheorem{lemma}{Lemma}

\definecolor{color0}{RGB}{216,27,96}
\definecolor{color1}{RGB}{30,136,229}
\definecolor{color2}{RGB}{255,193,7}
\definecolor{color3}{RGB}{0,77,64}
\definecolor{color4}{RGB}{36,158,209}

\acrodef{ltl}[LTL]{linear temporal logic}
\acrodef{mdp}[MDP]{Markov decision process}
\acrodef{scltl}[scLTL]{syntactically co-safe LTL}
\acrodef{dfa}[DFA]{deterministic finite-state automaton}
\acrodef{milp}[MILP]{mixed-integer linear programming}
\acrodef{dra}[DRA]{deterministic Rabin automaton}
\acrodef{ppo}[PPO]{Proximal Policy Optimization}
\acrodef{ddpg}[DDPG]{deep deterministic policy gradient}
\acrodef{dqn}[DQN]{deep Q-learning}
\acrodef{sac}[SAC]{soft actor-critic}
\acrodef{a2c}[A2C]{Advantage Actor Critic}
\acrodef{rl}[RL]{reinforcement learning} 

\begin{document}
\bstctlcite{BSTcontrol}

\title{Topology-Guided Modular Actor-Critic Learning for\\ Continuous Systems under Temporal Objectives}
\author{\IEEEauthorblockN{Lening Li\textsuperscript{1,*}~\orcidlink{0000-0002-8321-1254},
Zhentian Qian\textsuperscript{2}~\orcidlink{0000-0002-9826-6434},
Jianan Xia\textsuperscript{3},
Yawen Wang\textsuperscript{3},
Zhongjing Li\textsuperscript{3},\\
Qiren Geng\textsuperscript{3},
Huasheng Zhang\textsuperscript{3},
Liang Hu\textsuperscript{3},
Qishuang Li\textsuperscript{3},
Junqiang Lou\textsuperscript{1}~\orcidlink{0000-0002-8970-8472}}
\IEEEauthorblockA{\textsuperscript{1}\textit{Ningbo University}, Ningbo, China\\
\textsuperscript{2}\textit{Robotics Engineering Program, Worcester Polytechnic Institute}, Worcester, MA, USA\\
\textsuperscript{3}\textit{Ningbo Zhenhai Silver-Ball Bearing Co., Ltd.}, Ningbo, China\\
Emails: leningli@outlook.com, zqian@wpi.edu, loujunqiang@nbu.edu.cn,\\
\{xja, wyw, lizj, garry.geng, zhs, huliang, liqishuang\}@nzsbgroup.com\\
\textsuperscript{*}Corresponding author}
}

\maketitle

\begin{abstract}
We study formal policy synthesis for continuous-state stochastic systems under linear temporal logic specifications. The product of the system with the automaton of the specification has a hybrid state space with sparse rewards. We introduce a generalized optimal backup order, defined in reverse to a topological order over automaton states, that guides value backups and provably preserves optimality. We further present a model-free actor-critic algorithm whose policy evaluation solves a constrained optimization problem by the augmented Lagrangian method, yielding hyperparameter self-tuning, and prove its optimality and convergence in the tabular case. Since integer encodings of automaton states impose a spurious ordinal relationship on functions learned by one network, we dedicate a value and a policy network to each automaton state (modular learning). The algorithm matches or outperforms PPO, DQN, and A2C on CartPole, and on a Dubins car under a temporal specification the topological order and modular learning raise the success rate from $26.0\%$ to $71.5\%$.
\end{abstract}


\section{Introduction}
\label{sec:introduction}
Formal policy synthesis produces policies with provable probabilistic guarantees of satisfying a high-level specification, which can encode behaviors far richer than A-to-B planning (\eg, a low-fuel taxi that must avoid downtown and refuel before or after dropping off its passenger); such guarantees matter across safety-critical robotics and cyber-physical systems~\cite{belta_et_al:DR:2017:8281}. Two obstacles separate theory from deployment: the system model is often unavailable, and real systems are continuous-state, whereas most results address discrete systems.

We address both for continuous-state stochastic systems with specifications in \ac{ltl}~\cite{manna2012temporal}. Following the automaton-based route, we translate the specification into an automaton~\cite{duret2016spot}, form its product with the system, and reward reaching the accepting states, which in the hybrid product space are rarely reached, so rewards are sparse. Our remedy is structural: we generalize the optimal backup order of an \ac{mdp} to the product \ac{mdp} by backing up values in reverse to a \emph{topological order} over automaton states, which partitions product states into level sets learned level by level, so the system collects rewards whenever it enters a previously solved level; we prove that this preserves optimality.

We further propose a model-free actor-critic \ac{rl} algorithm compatible with this order. Unlike \ac{sac} and temporal-difference evaluation, its policy evaluation follows the linear programming solution of an \ac{mdp}~\cite{de2003linear}: with the mellowmax operator we form a constrained problem, impose its constraints only on sampled trajectories, eliminate transition probabilities through an unbiased estimate from sampled next states and rewards, and convert inequalities into equalities via the augmented Lagrangian method. Solving the result as a sequence of subproblems with fixed hyperparameters, updated between subproblems, grants hyperparameter self-tuning, and we prove convergence and optimality in the tabular case. Finally, since one network over the hybrid space implicitly ranks automaton states by their integer codes~\cite{yuan2019modular}, we assign each automaton state its own value and policy networks (\emph{modular learning}).

\paragraph*{Related work} \emph{Non-automaton-based} approaches encode \ac{ltl} as optimization constraints---an NP-complete \ac{milp} for nonlinear systems~\cite{wolff2014optimization} or satisfiability modulo convex programming for multi-robot planning~\cite{shoukry2017linear}---but require explicit models. \emph{Automaton-based} approaches~\cite{ding2011ltl, fu2014probably, hasanbeig2019reinforcement, hasanbeig2019certified, lavaei2020formal} solve the product with the task automaton by optimization or \ac{rl}: \cite{ding2011ltl} uses a \ac{dra} and a linear program with known transitions and discrete states; \cite{fu2014probably} learns a probably approximately correct \ac{mdp}, and~\cite{hasanbeig2019reinforcement} applies Q-learning to an on-the-fly product, both for discrete states; \cite{hasanbeig2019certified} gives the first model-free synthesis for continuous-state \acp{mdp} but inherits its policy evaluation from existing \ac{rl} and does not exploit the automaton structure to guide value backups. DIRL~\cite{jothimurugan2021compositional} combines model-based planning on an abstract graph of regions and sub-tasks with model-free \ac{rl} per edge, but still requires a dynamics model for planning.

\section{Preliminaries}
\label{sec:preliminary}
Let $\reals$ denote the reals, $2^X$ the power set of $X$, $\Sigma^\ast$ and $\Sigma^\omega$ the finite and infinite words over an alphabet $\Sigma$, and $\indicator(E)$ the indicator of event $E$. An \ac{ltl} formula over atomic propositions $\calAP$ is defined inductively as
$\varphi \coloneqq \top \mid p \mid \neg\varphi \mid \varphi_1 \land \varphi_2 \mid \Next \varphi \mid \varphi_1 {\until} \varphi_2$,
where $p \in \calAP$, $\Next \varphi$ (``next'') holds iff $\varphi$ holds at the next state, and $\varphi_1 \until \varphi_2$ (``until'') iff $\varphi_2$ eventually holds and $\varphi_1$ holds until then. The derived operators are $\Eventually \varphi = \top \until \varphi$ (eventually) and $\Always \varphi = \neg \Eventually \neg \varphi$ (always); the satisfaction relation $w \models \varphi$ for $w \in (2^\calAP)^\omega$ is standard~\cite{pnueli1989synthesis}.

We restrict specifications to \ac{scltl}~\cite{kupferman2001model}, whose formulas contain only $\Eventually$, $\Next$, and $\until$ in positive normal form~\cite{baier2008principles}; satisfaction then requires only a \emph{good prefix} $w \in \Sigma^\ast$ with $ww' \models \varphi$ for all $w' \in \Sigma^\omega$, and the good prefixes form the language of a \ac{dfa}.

A \ac{dfa} of $\varphi$ is a tuple $\dfa = \langle Q, \Sigma, \delta, \iota, F \rangle$ with a finite set $Q$ of states, alphabet $\Sigma = 2^\calAP$, deterministic transition function $\delta \colon Q \times \Sigma \to Q$, initial state $\iota$, and accepting states $F$. A word $w = \sigma_0 \ldots \sigma_n$ induces the run $q_0 q_1 \ldots q_{n+1}$ with $q_0 = \iota$ and $q_{i+1} = \delta(q_i, \sigma_i)$, and is accepted iff $q_{n+1} \in F$. We assume the \ac{dfa} is complete and minimal, with all rejecting behavior captured by a single sink state $q_\sink$ ($\delta(q_\sink, \sigma) = q_\sink$ for all $\sigma$).

\begin{example}
  \label{example:sequential_visiting}
  A \emph{sequential visiting} task requires a car to avoid obstacles ($O$) and either (a) visit $A$, then avoid $D$ until visiting $C$, or (b) visit $D$, then avoid $A$ until visiting $B$. With $\calAP = \{A, B, C, D, O\}$, the task is the \ac{scltl} formula $\varphi = \neg O \until (\varphi_1 \lor \varphi_2)$ with $\varphi_1 = A \land ((\neg D \land \neg O) \until C)$ and $\varphi_2 = D \land ((\neg A \land \neg O) \until B)$. Its \ac{dfa} is shown in Fig.~\ref{fig:dfa} (boxes are explained in Section~\ref{sec:synthesis}).
\end{example}

\begin{figure}[!t]
  \centering
  \resizebox{0.48\linewidth}{!}{
    \begin{tikzpicture}[->,>=stealth',shorten >=1pt,auto,node distance=4cm,scale=1,semithick, transform shape]
    \tikzstyle{every state}=[fill=black!10!white,minimum size=1cm,font=\Large]
    \draw[ultra thick,solid,color2] (-1,-1)		rectangle	(1, 1);
    \draw[ultra thick,dashed,color1] (3,-5)		rectangle	(5, 1);
    \draw[ultra thick,loosely dotted,color0]	(7,-1)  	rectangle	(9, 1);
    \draw[ultra thick,solid,color3]	(7,-5)  	rectangle	(9, -3);
    \node[] at (0, 1.6)	{\LARGE $X_0 \in \level_2$};
    \node[] at (4, 1.6)	{\LARGE $X_1 \in \level_1$};
    \node[] at (8, 1.6)	{\LARGE $X_2 \in \level_0$};
    \node[] at (8, -5.6)	{\LARGE $X_3 \in \level_0$};
    \node[initial,state] 	(0) 						{$q_0$};
    \node[state] 				(1) [right of=0]		{$q_1$};
    \node[state] 				(2) [below of=1]		{$q_2$};
    \node[state,accepting] 		(3) [right of=1]		{$q_3$};
    \node[state] 				(4) [below of=3]		{$q_4$};
    \path[->]
    (0) edge 				node {$\{D\}$}	    (1)
    (0) edge				node[left] {$\{A\}$}	    (2)
    (1) edge[bend right] 	node[left] {$\{A\}$}	    (2)
    (1) edge				node {$\{B\}$}		(3)
    (2) edge[bend right]	node[right] {$\{D\}$}	    (1)
    (2) edge 				node[right, near end] {$\{C\}$}		(3)
    (0) edge 	node[near end] {$\{O\}$}						(4)
    (1) edge 	node[near end] {$\{O\}$}  					(4)
    (2) edge 	node {$\{O\}$}  					(4)
    ;
    \end{tikzpicture}
  }
  \caption{\ac{dfa} of Example~\ref{example:sequential_visiting} (self-loops omitted); boxes are the maximal meta-modes $X_0 = \{q_0\}$, $X_1 = \{q_1, q_2\}$, $X_2 = \{q_3\}$, $X_3 = \{q_4\}$, and \eqref{eq:level_sets} gives $\level_0 = \{X_2, X_3\}$, $\level_1 = \{X_1\}$, $\level_2 = \{X_0\}$.}
  \label{fig:dfa}
\end{figure}

We model the system as a \emph{labeled \ac{mdp}} $\mdp = \langle S, A, P, s_0, L, \calAP \rangle$ with a set $S$ of states, a finite set $A$ of actions, transition probabilities $P(\cdot \mid s, a)$ over next states, a unique initial state $s_0$, atomic propositions $\calAP$, and a labeling function $L \colon S \to 2^{\calAP}$. A randomized policy $\pi \colon S \times A \to [0,1]$ ($\pi \in \Pi$) induces a distribution $\calD^\pi$ over \emph{paths} $\rho = s_0 a_0 s_1 a_1 \ldots$ with $a_t \sim \pi(\cdot \mid s_t)$, $s_{t+1} \sim P(\cdot \mid s_t, a_t)$; $\path^\pi$ denotes the paths under $\pi$ and $\calD^\pi(s)$ the path distribution from $s$. A finite path $\rho = s_0 a_0 \ldots s_N$ satisfies $\varphi$, $\rho \models \varphi$, iff $L(\rho) = L(s_0) \ldots L(s_N)$ is accepted by the \ac{dfa}.

\emph{MaxProb problem:} given a labeled \ac{mdp} $\mdp$ and an \ac{scltl} formula $\varphi$, synthesize a policy maximizing the probability of satisfying $\varphi$,
\begin{align}
  \pi^\ast = \argmax_{\pi \in \Pi} \Prob_{\rho \sim \calD^\pi}(\exists t \ge 0, \rho_t \models \varphi),
  \label{eq:maxprob}
\end{align}
where $\rho_t$ is the length-$t$ prefix of $\rho$.

\section{Automaton-based Formal Policy Synthesis}
\label{sec:synthesis}
\begin{definition}[Product MDP]
  \label{def:prodmdp}
  Given a labeled \ac{mdp} $\mdp$ and the \ac{dfa} $\dfa$ of $\varphi$, the product \ac{mdp} is $\prodmdp = \mdp \otimes \dfa = \langle Z, A, \prodtrans, z_0, \prodf \rangle$, where $Z = S \times Q$ is the set of product states $z = (s, q)$, with $q$ tracking progress toward satisfaction; $A$ is inherited from $\mdp$; $\prodtrans((s', q') \mid (s, q), a) = P(s' \mid s, a)\,\indicator(q' = \delta(q, L(s')))$; $z_0 = (s_0, \delta(\iota, L(s_0)))$; and $\prodf = S \times F$ is the set of final states, which we make absorbing by redefining $\prodtrans$ so that each final state transitions to itself with probability one.
\end{definition}

To solve the MaxProb problem~\eqref{eq:maxprob}, we define the reward $R \colon Z \times A \to \reals$, for each $z \in Z$ and $a \in A$, as
\begin{align}
  R(z, a) = \indicator(z \in {Z \setminus \prodf}) \sum_{z' \in Z} \prodtrans(z' \mid z, a) \indicator(z' \in \prodf),
  \label{eq:reward}
\end{align}
so that a reward is received only upon transiting into $\prodf$. The MaxProb problem then becomes optimal planning on $\prodmdp$ with value function $\val^\pi(z) = \Expect_{\rho \sim \calD^\pi(z)}[\sum_{t \ge 0} \gamma^t R(z_t, a_t)]$, $\gamma \in [0, 1)$, and optimal policy $\pi^\ast(z) = \argmax_{\pi} \val^\pi(z)$, where (abusing notation) $\pi \colon Z \times A \to [0,1]$. Reward~\eqref{eq:reward} is the probability of entering $\prodf$ at the next step; since $\prodf$ is absorbing, entry occurs at most once per path, so the undiscounted value equals the satisfaction probability, which the discounted value lower-bounds and recovers as $\gamma \to 1$.

\paragraph*{Hierarchical decomposition}
\label{sec:hierarchical_decomposition}
Reward~\eqref{eq:reward} is sparse: in a large or continuous state space the system rarely reaches $\prodf$. We address this by generalizing the optimal backup order of an \ac{mdp}~\cite{bertsekas2011dynamic} to the product \ac{mdp}. In an \ac{mdp}, state $s$ \emph{causally depends} on $s'$ if $P(s' \mid s, a) > 0$ for some $a$ and the Bellman equation\footnote{With the mellowmax operator $\mellowmax$ of equation~\eqref{eq:mellowmax_optimal} in this work.} makes $\val(s)$ depend on $\val(s')$~\cite{dai2011topological}; it is then more efficient to back up $s'$ before $s$, and if the \ac{mdp} is acyclic there exists an optimal backup order under which each state needs only one backup~\cite{bertsekas2011dynamic}.

In $\prodmdp$, $(s, q)$ causally depends on $(s', q')$, written $(s, q) \relation (s', q')$, if $\prodtrans((s', q') \mid (s, q), a) > 0$ for some $a \in A$. Computing this relation over a large or hybrid $Z$ is expensive; we observe instead that if $q' = \delta(q, \sigma)$ represents strictly more progress than $q$, the backup of $(s', q')$ should precede that of $(s, q)$ for any $s, s' \in S$. This motivates the \emph{guard set}.

\begin{definition}[Guard Set]
  \label{def:guard_set}
  For $q, q' \in Q$, the guard set is the set of \ac{mdp} states from which some action moves the automaton from $q$ to $q'$ with positive probability:
  \begin{align}
    \label{eq:guard_set}
    \guard(q, q', \mdp) = \{s \in S \mid \exists a \in A, \exists s' \in S, \nonumber \\
    P(s' \mid s, a) > 0 \land \delta(q, L(s')) = q'\}.
  \end{align}
\end{definition}

\begin{definition}[Causal Dependence on $Q$ and $\partition$]
  \label{def:causal_dep_Q}
  $q \relation q'$ if $\guard(q, q', \mdp) \neq \emptyset$. Let $\relation^+$ be the transitive closure of $\relation$ (on $Z$, $Q$, or $\partition$). $q_1, q_2$ are \emph{mutually} causally dependent, $q_1 \brelation q_2$, if $q_1 \relation^+ q_2$ and $q_2 \relation^+ q_1$. A \emph{meta-mode} $X \subseteq Q$ is a set of mutually dependent states ($\{q\}$ if $q$ depends mutually on no other state); a \emph{maximal meta-mode} also satisfies $q \not\brelation q'$ for all $q \in X$, $q' \notin X$. $\partition$ denotes the set of maximal meta-modes, and $X \relation X'$ if $q \relation q'$ for some $q \in X$, $q' \in X'$.
\end{definition}
Mutually dependent product states have no natural update order, so we update them together; the maximal meta-modes are the strongly connected components of $(Q, \relation)$.

\begin{lemma}
  \label{lm:partition}
  $\partition$ is a partition of $Q$.
\end{lemma}
\begin{proof}
  Every $q$ lies in a maximal meta-mode (the one containing the states it is mutually dependent on, or $\{q\}$). If distinct $X, X'$ shared $q$, every pair in $X \times X'$ would be mutually dependent via $q$, contradicting maximality.
\end{proof}

\begin{lemma}
  \label{lm:discrete_modes_casaul_dependence}
  If $X \relation^+ X'$ but not $X' \relation^+ X$, then for any $(s,q) \in S\times X$ and $(s',q')\in S\times X'$, either $(s,q) \relation^+ (s',q')$ and $(s',q') \not \relation^+ (s,q)$, or the two states are causally independent.
\end{lemma}
\begin{proof}
  It suffices to show $(s',q') \not\relation^+ (s,q)$. Otherwise there is $(s'', q'')$ with $(s',q') \relation^+ (s'',q'')$ and $(s'',q'') \relation (s,q)$; by Def.~\ref{def:guard_set}, $s'' \in \guard(q'', q, \mdp)$ and $q'' \relation q$, hence $q' \relation^+ q'' \relation q$ and $X' \relation^+ X$, a contradiction.
\end{proof}

Lemma~\ref{lm:discrete_modes_casaul_dependence} says that if $X \relation^+ X'$ and $X' \not\relation^+ X$, then $S \times X'$ should be updated before $S \times X$; it does not induce a \emph{total} order, since meta-modes can be causally independent. We resolve this by grouping meta-modes into \emph{level sets}.
\begin{definition}[Level Sets and Topological Order]
  \label{def:topological_order}
  Let $\level_0 = \{X \in \partition \mid (F \cup \{q_\sink\}) \cap X \ne \emptyset\}$ and, for $i \ge 1$, with $\bar\partition_i = \cup_{k < i} \level_k$,
  \begin{align}
    \level_i = \{X \in \partition \setminus \bar\partition_i \mid {} & \exists X' \in \level_{i-1}, X \relation X'; \label{eq:level_sets} \\
    & \forall X'' \in \partition \setminus (\bar\partition_i \cup \{X\}), X \not\relation X''\}, \nonumber
  \end{align}
  computed until $\level_{n+1} = \emptyset$; we write $q \in \level_i$ if $q$ lies in a meta-mode of $\level_i$. The topological order $\topo$ over $\{\level_i \mid 0 \le i \le n\}$ is $\level_j \topo \level_k$ iff $j = k+1$, \ie, $\level_n \topo \dots \topo \level_0$.
\end{definition}
Thus $\level_0$ holds the meta-modes containing accepting or sink states and $\level_i$ those depending only on lower levels (Fig.~\ref{fig:dfa}).

\begin{theorem}[Generalized Optimal Backup Order]
  \label{theorem:topological_order}
  Updating the value function of each level set in reverse to $\topo$ finds the optimal value function of each level set with one backup per level set, \ie, one call to an optimal planner restricted to that level set.
\end{theorem}
\begin{proof}
  By induction on $i$; the base case is planning on $\level_0$. If all $\level_k$, $k < i$, are optimal, then $\val(s,q)$, $q \in \level_i$, depends only on $\{\val(s',q') \mid (s,q) \relation (s',q')\}$, and by Lemma~\ref{lm:discrete_modes_casaul_dependence} and~\eqref{eq:level_sets} every such $q'$ lies in some $\level_k$, $k \le i$: already optimal ($k < i$) or solved jointly ($k = i$). Hence level $i$, once converged, is never changed by higher levels.
\end{proof}

\emph{Topology-guided value learning} thus sets $\val = 0$ on $S \times \level_0$ (final and sink states only) and, for $i = 1, \dots, n$, calls any optimal planner on $S \times \level_i$ given the solved lower levels; Section~\ref{sec:SRL} provides that planner.

\section{Policy Synthesis on Each Level Set}
\label{sec:SRL}
With $\qval(z, a) = R(z, a) + \gamma \Expect_{z' \sim \prodtrans(\cdot \mid z, a)}[\val(z')]$ and temperature $\tau > 0$, the optimal mellowmax operator is
\begin{align}
  \mellowmax \val(z) = \tau \log \textstyle\sum_{a \in A}\exp\{\qval(z, a) / \tau\}, \quad z \in Z. \label{eq:mellowmax_optimal}
\end{align} Optimality is henceforth soft (entropy-regularized); as $\tau \to 0$, $\mellowmax$ recovers $\max$ and standard optimality. Given optimal values $\val^\ast$ on $S \times \level_k$ for all $k < i$, the problem for level $i$, in the spirit of the linear programming formulation~\cite{de2003linear}, is
\begin{align}
  \label{eq:topo_LTL}
  \min_{\val} \; & \sum_{(s, q) \in S \times \level_i} c(s, q)\val(s, q) \nonumber \\
  \mbox{s.t.} \; & \mellowmax \val(s, q) \leq \val(s, q), \; \forall (s, q) \in S \times \level_i,
\end{align}
where $c(s, q) > 0$ are \emph{state-relevance weights} and $\val(s', q')$ is a variable if $q' \in \level_i$ and a known constant otherwise. Its solution is $\val^\ast$ on $S \times \level_i$, $\qval^\ast$ follows by one Bellman backup, and $\pi^\ast(a \mid (s, q)) = \exp\{(\qval^\ast((s, q), a) - \val^\ast(s, q)) / \tau\}$. Problem~\eqref{eq:topo_LTL}, however, has one constraint per state and is intractable when $Z$ is hybrid.

\paragraph*{Sequential actor-critic algorithm}
We therefore approximate the value and policy on level $i$ by neural networks $\val_\theta(s, q)$, $\pi_\phi(\cdot \mid s, q)$ and, since $\prodmdp$ is an \ac{mdp} augmented with an automaton state, present the algorithm in the plain \ac{mdp} setting. Unlike a classic actor-critic, ours evaluates policies through a constrained optimization that enables hyperparameter self-tuning, alternating policy evaluation and improvement.

\paragraph{Policy Evaluation}
Define the mellowmax operator of a policy $\pi$ as $\mellowmax^\pi \val(s) = \sum_{a \in A} \pi(a \mid s)(\qval(s, a) - \tau \log \pi(a \mid s))$. Analogously to the linear programming characterization of policy evaluation~\cite{de2003linear}, $\val^\pi$ solves
\begin{align}
  \min_{\val} \sum_{s \in S} c(s)\val(s) \quad \mbox{s.t.} \quad \val(s) \ge \mellowmax^\pi \val(s), \; \forall s \in S. \label{prob:opt_prob}
\end{align}
Paralleling Lemma 1 of~\cite{de2003linear} with $\mellowmax^\pi$ in place of the Bellman operator, $\val$ solves~\eqref{prob:opt_prob} iff it solves $\min_{\val} \norm{\val^\pi - \val}_{1, c}$ under the same constraints. Let $c(s) = \pr^\pi(s)$, the (unnormalized) state occupancy measure of $\pi$; since $\sum_{s} \pr^\pi(s) f(s) = \Expect_{\rho \sim \calD^\pi}[\sum_{t \ge 0} f(s_t)]$ for any $f$, the objective of~\eqref{prob:opt_prob} becomes $\Expect_{\rho \sim \calD^\pi}[\sum_{t \ge 0} \val(s_t)]$. Let $g(s) = \mellowmax^\pi \val(s) - \val(s) = \sum_{a} \pi(a \mid s)\big(R(s, a) + \gamma \Expect_{s'}[\val(s')] - \tau \log \pi(a \mid s)\big) - \val(s)$ and let $h \colon \reals \to \reals$ be continuous with $h(x) = 0$ for $x \le 0$ and $h(x) > 0$ for $x > 0$, \eg, $h(x) = (\max\{x, 0\})^2$, so that the inequality constraint becomes $h(g(s)) = 0$. Because $g$ requires the transition probabilities, we replace $R(s, a) + \gamma \Expect_{s'}[\val(s')]$ by the unbiased estimate $R(s, a) + \gamma \val(s')$~\cite{sutton2018reinforcement} with $s' \sim P(\cdot \mid s, a)$ a sampled next state for each $a$:
\begin{align}
  \tilde{g}(s) = {} & \textstyle\sum_{a \in A} \pi(a \mid s)\big(R(s, a) \nonumber \\ & \qquad + \gamma \val(s') - \tau \log \pi(a \mid s)\big) - \val(s). \label{eq:g}
\end{align} Although $\tilde{g}$ is unbiased for $g$, $\Expect[h(\tilde{g}(s))] \ne h(g(s))$ for nonlinear $h$: under stochastic transitions the constraint also penalizes the variance of $\tilde{g}$; the substitution is exact for deterministic transitions. Since a continuous $S$ yields infinitely many constraints, we constrain only the states visited by $\pi$, which agrees with~\eqref{prob:opt_prob} on the support of $\pr^\pi$:
\begin{align}
  \min_{\val} \Expect_{\rho \sim \calD^\pi}\big[\textstyle\sum_{t \ge 0} \val(s_t)\big] \quad \mbox{s.t.} \quad \Expect_{\rho \sim \calD^\pi}[h(\tilde{g}(s_t))] = 0. \label{prob:translated_approximate_problem}
\end{align}
The augmented Lagrangian of a path $\rho = s_0 a_0 s_1 a_1 \dots$ is
\begin{align}
  \Lagrange(\rho, \vec{\lambda}, \vec{\nu}) = \textstyle\sum_{t\ge 0}\big[\val(s_t) + \lambda_t h(\tilde{g}(s_t)) + \tfrac{\nu_t}{2}h(\tilde{g}(s_t))^2\big], \label{eq:lagrange}
\end{align}
with $\vec{\lambda} = [\lambda_0, \lambda_1, \dots]^\intercal$ and $\vec{\nu} = [\nu_0, \nu_1, \dots]^\intercal$, and problem~\eqref{prob:translated_approximate_problem} becomes $\min_{\val} \Expect_{\rho \sim \calD^\pi}[\Lagrange(\rho, \vec{\lambda}, \vec{\nu})]$. We solve it \emph{sequentially}, as a sequence of subproblems with fixed multipliers, the $m$-th being $\min_{\val} \Expect_{\rho \sim \calD^\pi}[\Lagrange(\rho, \vec{\lambda}_m, \vec{\nu}_m)]$.

\paragraph{Policy Improvement}
Policy improvement minimizes the soft consistency error~\cite{nachum2017bridging}, $J^\pi = \Expect_{\rho \sim \calD^\pi}[\tfrac{1}{2}C(\rho)^2]$, where for a finite path $\rho = s_0 a_0 \cdots s_T$
\begin{align}
  C(\rho) = {} & - \val^{\pi}(s_0) + \gamma^T \val^{\pi}(s_{T}) \label{eq:T_step_soft_consistency} \\
  & + \textstyle\sum_{t = 0}^{T-1} \gamma^t (R(s_t, a_t) - \tau \log \pi(a_t \mid s_t)). \nonumber
\end{align}
Since $\val(s) = R(s, a) + \gamma \Expect_{s' \sim P(\cdot \mid s, a)}[\val(s')] - \tau \log \pi(a \mid s)$ for all $(s, a)$ implies $\val = \val^\ast$ and $\pi = \pi^\ast$~\cite{nachum2017bridging}, $J^\pi = 0$ implies $\pi = \pi^\ast$.

\begin{theorem}[Policy Iteration]
  \label{thm:policy_iteration}
  Repeated application of exact policy evaluation~\eqref{prob:opt_prob} and policy improvement $\min_\pi J^\pi$ from any $\pi \in \Pi$ converges to the soft-optimal policy $\pi^\ast$ with $\qval^{\pi^\ast}(s, a) \ge \qval^{\pi}(s, a)$ for all $\pi \in \Pi$ and $(s, a) \in S \times A$.
\end{theorem}
\begin{proof}
  Parallels Theorem 1 of~\cite{haarnoja2018soft}.
\end{proof}

\begin{algorithm}[!t]
  \caption{Sequential Actor-Critic Algorithm}
  \label{alg:sac}
  \footnotesize
  \begin{algorithmic}[1]
    \renewcommand{\algorithmicrequire}{\textbf{Input:}}
    \renewcommand{\algorithmicensure}{\textbf{Output:}}
    \REQUIRE $M$, $N$, $\beta$, $\epsilon$, $\eta$, $K$, $\vec{\lambda}_0$, $\vec{\nu}_0$; environment $\env$.
    \ENSURE $\val_{\theta}, \pi_{\phi}$. \textit{Init:} random $\theta, \phi$; empty replay buffer $\mathcal{B}$; $m = 0$.
    \WHILE{$m < M$}
    \STATE{$\mathsf{viol}_m \leftarrow \mathsf{Violation}(\theta, \phi)$}
    \FOR{$n = 1$ to $N$}
    \STATE{Sample a path $\rho$ under $\pi_{\phi}$ from $\env$; add $\rho$ to $\mathcal{B}$.}
    \STATE{Sample $K$ trajectories $\rho_1, \dots, \rho_K$ from $\mathcal{B}$.}
    \STATE{$\theta \leftarrow \theta - \eta \nabla_{\theta} \frac{1}{K}\sum_{k =1}^{K} \Lagrange^{\phi}_{\theta}(\rho_k, \vec{\lambda}_m, \vec{\nu}_m)$}
    \STATE{$\phi \leftarrow \phi - \eta \frac{1}{K}\sum_{k=1}^{K} C_{\theta}^{\phi}(\rho_k) \sum_{t=0}^{\abs{\rho_k}-2} \gamma^t \nabla_{\phi} \log \pi_{\phi}(a_t \mid s_t)$}
    \ENDFOR
    \STATE{$\mathsf{viol}_{m+1} \leftarrow \mathsf{Violation}(\theta, \phi)$}
    \STATE{$\vec{\lambda}_{m+1} \leftarrow \vec{\lambda}_m + \vec{\nu}_m \, \mathsf{viol}_{m+1}$}
    \STATE{$\vec{\nu}_{m+1} \leftarrow \beta \vec{\nu}_m$ if $\mathsf{viol}_{m+1} > \epsilon \, \mathsf{viol}_{m}$, else $\vec{\nu}_m$}
    \STATE{$m \leftarrow m+1$}
    \ENDWHILE
    \RETURN $\val_{\theta}, \pi_{\phi}$
  \end{algorithmic}
\end{algorithm}

\paragraph{Practical Algorithm}
For a continuous-state \ac{mdp}, let $\tilde{g}_{\theta}^{\phi}$, $\Lagrange_{\theta}^{\phi}$, $C_{\theta}^{\phi}$ denote~\eqref{eq:g}, \eqref{eq:lagrange}, \eqref{eq:T_step_soft_consistency} with $\val, \pi$ replaced by $\val_\theta, \pi_\phi$. The $m$-th subproblem becomes $\min_\theta \Expect_{\rho \sim \calD^{\pi_\phi}}[\Lagrange^\phi_\theta(\rho, \vec{\lambda}_m, \vec{\nu}_m)]$ and policy improvement $\min_\phi J^\phi_\theta = \Expect_{\rho \sim \calD^{\pi_\phi}}[\tfrac{1}{2}C^\phi_\theta(\rho)^2]$, whose gradient parallels path consistency learning~\cite{nachum2017bridging}; both are minimized by gradient descent with expectations estimated from $K$ replay-buffer trajectories (Algorithm~\ref{alg:sac}); following~\cite{haarnoja2018soft}, the critic comprises two networks whose minimum is used (clipped double critic). Each subproblem runs $N$ iterations with fixed multipliers, each sampling one trajectory and taking one evaluation and one improvement step; the multipliers are then updated from the violation $\mathsf{Violation}(\theta, \phi) = \frac{1}{K}\sum_{k=1}^{K}\sum_{t=0}^{\abs{\rho_k}-2} h(\tilde{g}^{\phi}_{\theta}(s_t))$ over $K$ buffered trajectories, with penalty growth rate $\beta$ and threshold $\epsilon$. In practice all $\lambda_t$ (resp.\ $\nu_t$) share one value $\lambda$ (resp.\ $\nu$).

\paragraph{Modular Learning}
\label{sec:modular}
A single network fed the integer code of the automaton state implicitly ranks automaton states by those integers, which empirically hampers training. We therefore learn, for each $q \in \level_i$, separate networks $\val_{\theta_q}(s)$ and $\pi_{\phi_q}(\cdot \mid s)$---$2\abs{\level_i}$ per level---switching networks as the automaton state changes~\cite{yuan2019modular}; this does not affect the topological order.

\begin{figure}[!t]
  \centering
  \begin{subfigure}[b]{0.46\linewidth}\centering\includegraphics[width=\linewidth]{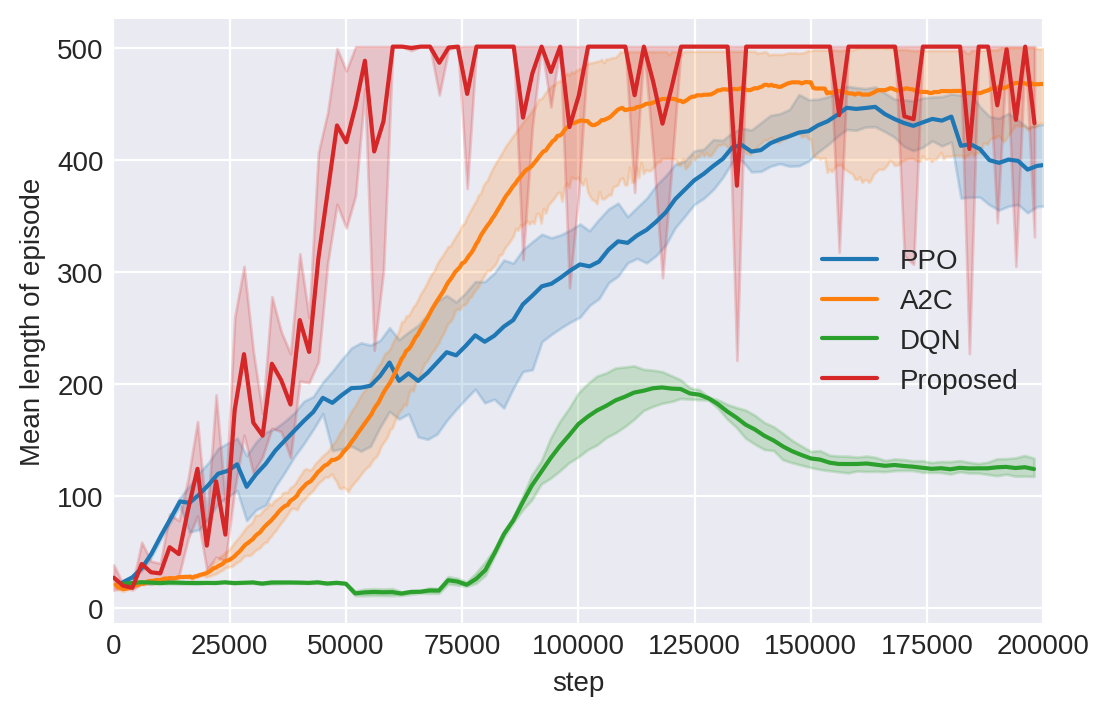}\caption{}\end{subfigure}\hspace{0.02\linewidth}
  \begin{subfigure}[b]{0.46\linewidth}\centering\includegraphics[width=\linewidth]{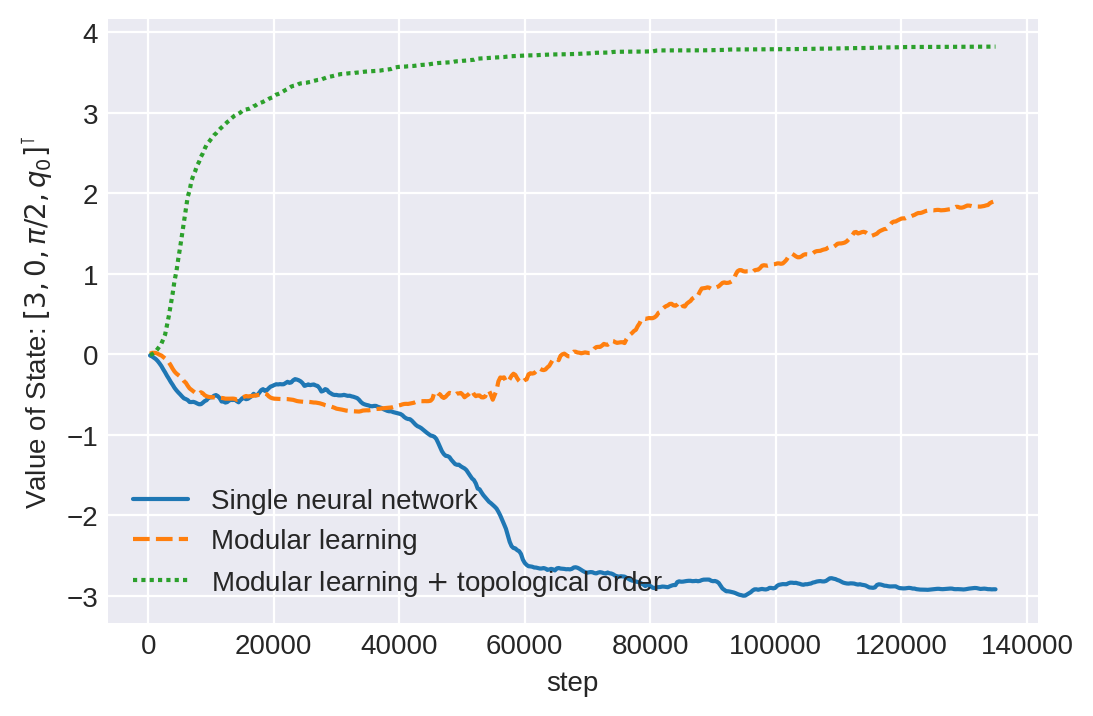}\caption{}\end{subfigure}
  \caption{(a) Mean episode length on CartPole-v1. (b) Initial-state value $\val(z_0)$, $z_0 = [3, 0, \pi/2, q_0]^\intercal$, for the three Dubins car variants.}
  \label{fig:results}
\end{figure}
\section{Case Study}
\label{sec:case_study}
We evaluate the \ac{rl} algorithm on CartPole against \ac{ppo}~\cite{schulman2017proximal}, \ac{dqn}~\cite{mnih2013playing}, and \ac{a2c}~\cite{mnih2016asynchronous}, then demonstrate the full framework on a Traxxas Slash $4 \times 4$ RC car. Both tasks use $\eta = 3 \cdot 10^{-4}$, $\gamma = 0.99$, two hidden layers of $256$ units, replay buffer $10^4$, $T = 10$, $\beta = 2$, $\epsilon = 0.9$, $\nu = 10^5$; CartPole uses $\tau = 1$, $\lambda = 10^4$, $M = 4$, $N = 2500$, $K = 10$; the sequential visiting task uses $\tau = 0.5$, $\lambda = 10^3$, $M = 3$, $N = 1500$, $K = 5$, decaying $\eta$ by $0.5$ every $10^3$ steps. Experiments run on an AMD Ryzen 9 5900X, 32~GB RAM, and an NVIDIA RTX 3060.

\paragraph*{Classic control: proposed method versus baselines}
\label{sec:cartpole}
In CartPole-v1~\cite{brockman2016openai}, two actions push a cart to balance a pole on a passive joint; an episode ends when the pole exceeds $12$ degrees from vertical, the cart moves more than $2.4$ units from the center, or the length reaches $500$, with reward $1$ per step. We run our algorithm and the baselines five times each with random seeds. Fig.~\ref{fig:results}(a) shows that our algorithm matches or exceeds \ac{ppo}, \ac{a2c}, and \ac{dqn}. The initial-state value converges after $6 \times 10^4$ steps; at $5 \times 10^4$ steps the actor and critic losses and the constraint violation approach zero as the episode length nears $500$. Training takes about $4$ hours, dominated by the rollouts used to evaluate the mean episode length (which converges to $500$ although training uses only length-$T$ segments).

\paragraph*{Robot motion planning with a high-level specification}
\label{sec:robot_motion_planning}
We apply the framework to the sequential visiting task of Example~\ref{example:sequential_visiting} in a planar workspace with regions $A$--$D$ and rectangular obstacles. The car is a Dubins car, $\dot{x} = v \cos \theta$, $\dot{y} = v \sin \theta$, $\dot{\theta} = (v/\ell) \tan \delta$, with position $[x, y]$, heading $\theta$, constant speed $v = 0.3$~m/s, wheelbase $\ell = 0.32$~m, and steering angle $\delta \in \{-2\pi/15, 0, 2\pi/15\}$ (turn rate $\approx \pm 0.42$~rad/s), starting from $[3, 0, \pi/2]^\intercal$. With $\vec{z} = [x, y, \theta]^\intercal$, the discrete-time evolution is $\vec{z}_{t+\Delta t} = \vec{z}_t + \dot{\vec{z}}_t \Delta t + \Delta\vec{z}_{\mathrm{noise}}$ with $\Delta t = 1$~s and zero-mean Gaussian noise of standard deviation $10^{-2}$ per component.

A reward of $1$ upon satisfaction is easily overwhelmed by the policy entropy under the mellowmax operator, so we use $10$ upon completing the specification and $-1$ upon leaving the workspace or hitting an obstacle. To further mitigate sparsity we add the shaped reward $r(\vec{z}, \vec{z}_{\subgoal}) = 5\,[\dot{x}, \dot{y}]^\intercal \cdot \vec{d}/\norm{\vec{d}}$ with $\vec{d} = [x_{\subgoal}-x, y_{\subgoal}-y]^\intercal$, where the subgoal $\vec{z}_{\subgoal}$ is $[1.25, 1.25]^\intercal$, $[4.25, 4.25]^\intercal$, and $[4.25, 1.25]^\intercal$ for $q_0$, $q_1$, and $q_2$ (none for $q_3$, $q_4$, where the specification is satisfied or violated). This shaping is a practical heuristic: the guarantee of Section~\ref{sec:synthesis} applies to reward~\eqref{eq:reward}, and shaping trades that exact correspondence for sample efficiency. The learned policy also completes the task from initial states unseen in training, \eg, $[3, 2, -\pi]^\intercal$.

\begin{table}[!t]
  \centering
  \caption{Success Rates over $200$ Simulations (Wilson $95\%$ CI) and Training Time}
  \label{tab:success_rate}
  \small\setlength{\tabcolsep}{4pt}
  \begin{tabular}{l|ccc}
    \hline
    Variant                                & Success Rate & $95\%$ CI & Time \\ \hline
    Single neural network                  & $26.0\%$     & $[20.4\%, 32.5\%]$ & $17$~min \\
    Modular learning                       & $49.0\%$     & $[42.2\%, 55.9\%]$ & $23$~min \\
    Modular $+$ topo.\ order               & $71.5\%$     & $[64.9\%, 77.3\%]$ & $53$~min \\
    \hline
  \end{tabular}
\end{table}

To isolate the two mechanisms we compare a single network over the product \ac{mdp} (input: car state and integer automaton state), modular learning (input: car state), and modular learning with the topological order. Table~\ref{tab:success_rate} reports success rates over $200$ simulations and Fig.~\ref{fig:results}(b) the initial-state value. Modular learning raises the success rate substantially, indicating that breaking the ordinal relationship yields better approximations, and the topological order raises it further for the same training steps while attaining the highest value and fastest convergence; the Wilson $95\%$ intervals are pairwise disjoint, so the gaps exceed sampling noise. The topological-order variant should theoretically perform at least as well as the one without; the observed gain confirms that the order guides value backups, at about twice the training time since both $\level_1$ and $\level_2$ are learned. A video\footnote{\url{https://tinyurl.com/3dcysrux}} shows the controller satisfying the specification on the RC car (localized with Tag36h11 AprilTags~\cite{olson2011apriltag}).

\paragraph*{Gap between the formal objective and demonstrated performance}
The MaxProb guarantee attaches to reward~\eqref{eq:reward} and exact values, whereas the experiments use reward amplification, shaping, and neural approximation; the success rates are thus empirical estimates, not certified probabilities. The gap to $100\%$ reflects shaping and approximation error and the finite training budget, not a limitation of the framework, whose optimality results (Theorems~\ref{theorem:topological_order} and~\ref{thm:policy_iteration}) hold for the unshaped objective; certified bounds for neural policies are future work.

\section{Conclusion}
\label{sec:conclusion}
We proposed a formal policy synthesis framework for continuous-state stochastic systems under high-level specifications: a topological order that overcomes reward sparsity while provably preserving optimality, a compatible sequential actor-critic \ac{rl} algorithm with tabular guarantees, and modular learning that breaks the ordinal relationship among automaton states. The algorithm is competitive with standard baselines on CartPole, and on a Dubins car under a temporal specification the two mechanisms raise the success rate from $26.0\%$ to $71.5\%$.

\bibliographystyle{IEEEtran}
\bibliography{references}

\end{document}